\documentclass[10pt,twocolumn,letterpaper]{article}

\usepackage{cvpr}

\usepackage{times}
\usepackage{epsfig}
\usepackage{graphicx}
\usepackage{amsmath}
\usepackage{amssymb}
\usepackage{nicefrac}
\usepackage{tabularx}
\usepackage{booktabs}
\usepackage{color, colortbl}
\usepackage{multirow}
\usepackage{makecell}
\usepackage{enumitem}
\usepackage{color,soul}
\usepackage{subcaption}

\usepackage{floatrow}
\usepackage{caption}
\newfloatcommand{tabbox}{table}[][\FBwidth]

\definecolor{Gray}{gray}{0.9}

\usepackage[pagebackref=true,breaklinks=true,bookmarks=false]{hyperref}

\cvprfinalcopy

\usepackage[pagebackref=true,breaklinks=true,bookmarks=false]{hyperref}

\begin{document}


\title{Mixed supervision for surface-defect detection:\\from weakly to fully supervised learning}

\author{Jakob Bo\v{z}i\v{c}, Domen Tabernik and Danijel Sko\v{c}aj\\
University of Ljubljana, Faculty of Computer and Information Science\\
Ve\v{c}na pot 113, 1000 Ljubljana\\
{\tt\small jakob.bozic@fri.uni-lj.si, domen.tabernik@fri.uni-lj.si}, \\ 
{\tt\small danijel.skocaj@fri.uni-lj.si}
}

\maketitle

\begin{abstract}
Deep-learning methods have recently started being employed for addressing surface-defect detection problems in industrial quality control. However, with a large amount of data needed for learning, often requiring high-precision labels, many industrial problems cannot be easily solved, or the cost of the solutions would significantly increase due to the annotation requirements. In this work, we relax heavy requirements of fully supervised learning methods and reduce the need for highly detailed annotations. By proposing a deep-learning architecture, we explore the use of annotations of different details ranging from weak (image-level) labels through mixed supervision to full (pixel-level) annotations on the task of surface-defect detection. The proposed end-to-end architecture is composed of two sub-networks yielding defect segmentation and classification results. The proposed method is evaluated on several datasets for industrial quality inspection: KolektorSDD, DAGM and Severstal Steel Defect. We also present a new dataset termed KolektorSDD2 with over 3000 images containing several types of defects, obtained while addressing a real-world industrial problem. We demonstrate state-of-the-art results on all four datasets. The proposed method outperforms all related approaches in fully supervised settings and also outperforms weakly-supervised methods when only image-level labels are available. We also show that mixed supervision with only a handful of fully annotated samples added to weakly labelled training images can result in performance comparable to the fully supervised model's performance but at a significantly lower annotation cost.
\end{abstract}
\maketitle


\section{Introduction}

Surface-quality inspection of production items is an important part of industrial production processes. Traditionally, classical machine-vision methods have been applied to automate visual quality inspection processes, however, with the introduction of Industry 4.0 paradigm, deep-learning-based algorithms have started being employed~\cite{Onchis2021,Yang2020,Tabernik2019JIM,Yu2019,Lin2018, Weimer2016}. Large capacity for complex features and easy adaptation to different products and defects without explicit feature hand-engineering made deep-learning models well suited for industrial applications. However, an essential aspect of the deep-learning approaches is the need for a large amount of annotated data that can often be difficult to obtain in an industrial setting. In particular, the data acquisition is often constrained by the insufficient availability of the defective samples and limitations in the image labeling process. In this paper, we explore how these limitations can be addressed with weakly and fully supervised learning combined into a unified approach of mixed supervision for industrial surface-defect detection.

In large-scale production, the quantity of the items is not the main issue but the ratio between non-defective and defective samples is by design heavily skewed towards the defect-free items. Often, anomalous items are a hundred, a thousand, or a even million-fold less likely to occur than the non-anomalous ones, leading to a limited set of defective items for training. Certain defect types can also be very rare with only a handful of samples available. This represents a major problem for supervised deep-learning methods that aim to learn specific characteristics of the defects from a large set of annotated samples.

The annotations themselves present an additional problem as they need to be detailed enough to differentiate the anomalous image regions from the regular ones properly. Pixel-level annotations are needed, but in practice, they are often difficult to produce. This is particularly problematic when it is challenging to explicitly define the boundaries between the defects and the regular surface appearance. Moreover, creating accurate pixel-level annotations is tiresome and costly to produce even when the boundaries are clear. Therefore, it is essential to minimize the labelling effort by decreasing the required amount of annotations and reducing the expected labels' precision.

\begin{figure}
    \centering
    \includegraphics[width=\columnwidth]{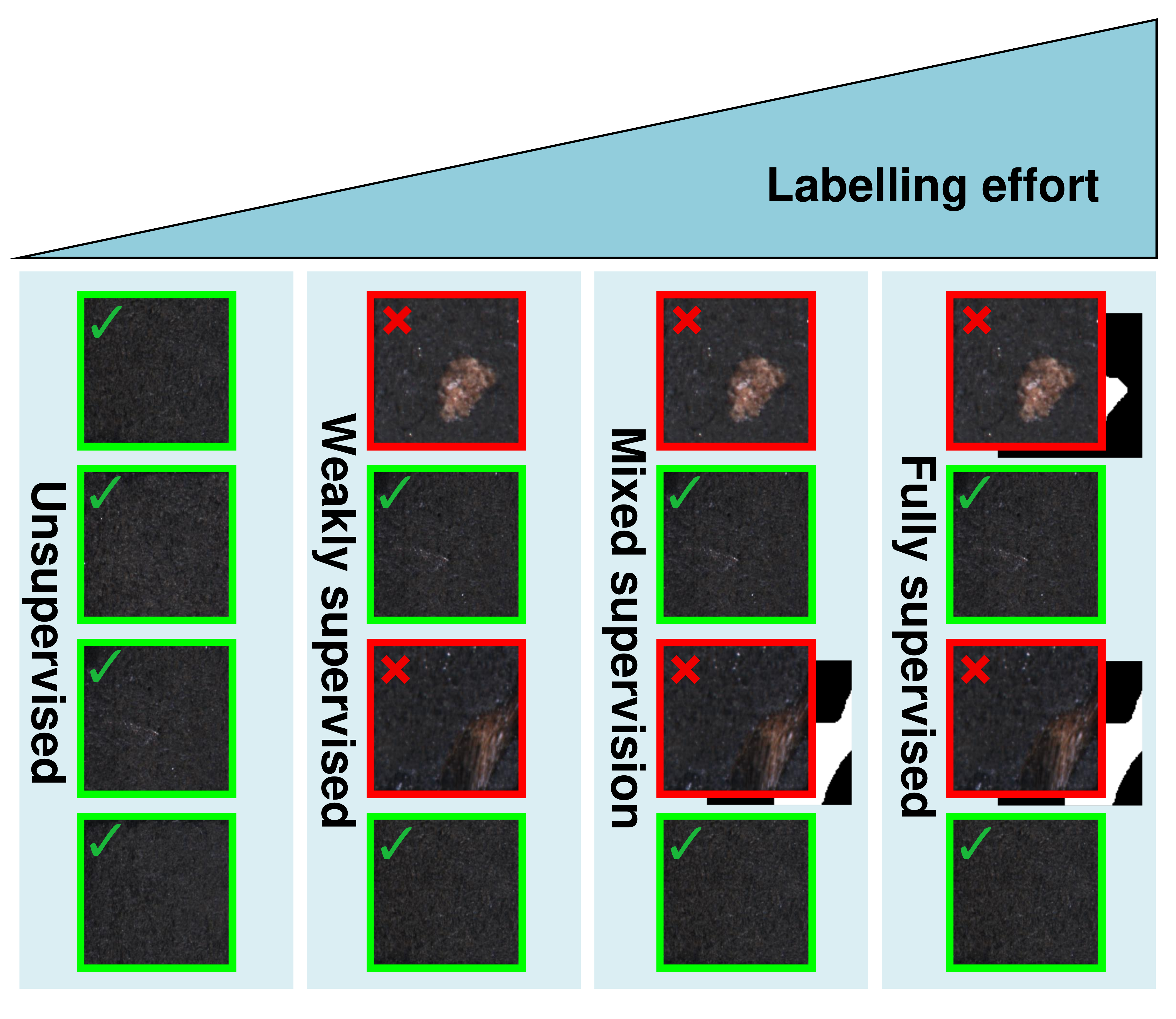}
    \caption{Visualization of several types of supervision and their required labelling effort.} 
    \label{fig:supervision-types}
\end{figure}

Various unsupervised~\cite{Staar2018,uninformedStudents,Zavrtanik2020} and weakly supervised~\cite{Lin2018,CADN,Zhu2019} deep-learning approaches have been developed to reduce the need for a large amount of annotated data. The former are designed to train the models using defect-free images only, while the latter utilise weak labels and do not require pixel-level annotations. 
Although both approaches significantly reduce the cost of acquiring the annotated data, they significantly underperform in comparison with fully supervised methods on the defect detection task. Moreover, for many industrial problems, a small amount of fully annotated data is often available and can be used for training the models to improve the performance. This can result in a mixed supervision mode with some fully labeled samples and a number of weakly labeled ones as depicted in Fig.~\ref{fig:supervision-types}.

Unsupervised and weakly supervised methods are not able to utilize these available data. Mixed supervision has been applied on other computer vision tasks, such as image segmentation~\cite{Souly2017,Mlynarski2019}, however, it has not yet been considered for industrial surface-defect detection.

In this work, we focus on a deep-learning approach suitable for industrial quality-control problems with varying availability of the defective samples and their annotations. In particular, we propose a deep-learning model for anomaly detection trained with weak supervision at the image-level labels, while at the same time utilising full supervision at the pixel-level labels when available. The proposed network can therefore work in a mixed supervision mode, fully exploiting the data and annotations available. By changing the amount and details of the required labels, the approach provides an option to find the most suitable trade-off between the annotation cost and classification performance. The mixed supervision is realised by implementing an end-to-end architecture with two sub-networks that are being simultaneously trained utilizing pixel-level annotations in the first sub-network and image-level annotations in the second sub-network.
While pixel-level labels can be utilized during the training to further increase the performance of the proposed method, our primary goal is not to segment the defects, but rather to identify images containing defective surfaces. In addition, we also account for spatial uncertainty in coarse region-based annotations to further reduce the annotation cost.

We performed an extensive evaluation of the proposed approach on several industrial problems. We demonstrate the performance on DAGM~\cite{Weimer2016}, KolektorSDD~\cite{Tabernik2019JIM} and Severstal Steel~\cite{SeverstalSteel2019} datasets. Due to the lack of a real-world, unsaturated, large, and well-annotated surface-defect dataset, we also compiled and made publicly available a novel dataset termed \textit{KolektorSDD2}
. It is based on a practical, real-world example, consisting of over 3000 images containing several defect types of various difficulty. Images have been carefully annotated to facilitate accurate benchmarking of surface-defect-detection methods.

The proposed model outperforms related unsupervised and weakly supervised approaches when only weakly labeled annotations are available, while in the fully supervised scenario it also outperforms all other related defect-detection methods.
Moreover, the proposed model demonstrates a significant performance improvement when only a few fully annotated images are added in mixed supervision. This often results in performance comparable to the results of fully supervised models but with a significantly reduced annotation cost.

The remainder of this paper is structured as follows: In Section~\ref{sec:related-work}, we present the related work, followed by the description of the proposed network with mixed supervision for surface-defect detection in Section~\ref{sec:method}.
We present the details of the experimental setup and evaluation datasets in Section~\ref{sec:experiment-setup}, while we present detailed evaluation results in Section~\ref{sec:results}. We conclude with a discussion in Section~\ref{sec:conclusion}.

\section{Related work} \label{sec:related-work}

\paragraph{Fully supervised defect detection} Several related works explored the use of deep-learning for industrial anomaly detection and categorisation~\cite{Onchis2021, Lin2020, Huang2020, Yu2019, Tabernik2019JIM, Racki2018,Lin2018, Wang2018c, Kim2017_DAGM}, including the early work of Masci et al.~\cite{Masci2012} using a shallow network for steel defect classification, and a more comprehensive study of a modern deep network architecture by Weimer et al.~\cite{Weimer2016}. From recent work, Kim et al.~\cite{Kim2017_DAGM} used a VGG16 architecture pre-trained on general images for optical inspection of surfaces, while Wang et al.~\cite{Wang2018c} applied a custom 11-layer network for the same task. Ra\v{c}ki et al.~\cite{Racki2018} further proposed to improve the efficiency of patch-based processing from~\cite{Weimer2016} with a fully convolutional architecture and proposed a two-stage network architecture with a segmentation net for pixel-wise localization of the error and a classification network for per-image defect detection. 
In our more recent work~\cite{Tabernik2019JIM}, we  performed an extensive study of the two-stage architecture with several additional improvements and showed the state-of-the-art results that outperformed others such as U-Net~\cite{Ronneberger2015} and DeepLabv3~\cite{Chen2017} on a real-world case of anomaly detection problem. We also extended this work and presented an end-to-end learning method for the two-stage approach~\cite{bozic2020}, however still in a fully-supervised regime, without considering the task in the context of mixed or weakly supervised learning.
Dong et al.~\cite{Dong2020} also used the U-Net architecture but combined it with SVM for classification and random forests for detection. Other recent approaches also explored lightweight networks~\cite{Yang2020, Lin2020, Huang2020, Liu2020a}. Lin et al.~\cite{Lin2020} used a compact multi-scale cascade CNN termed MobileNet-v2-dense for surface-defect detection while Huang et al.~\cite{Huang2020} proposed an even more lightweight network using atrous spatial pyramid pooling (ASPP) and depthwise separable convolution.

\paragraph{Unsupervised learning}

In unsupervised learning, annotations are not needed (and are not taken into account even when available) and features are learned from either reconstruction objective~\cite{Kingma2014,Chen2017b}, adversarial loss~\cite{Goodfellow2014} or similar self-supervised objective~\cite{Croitoru2017,Wang2017,Zhang2016a}.
In unsupervised anomaly detection solutions, the models are usually trained considering only non-anomalous image by applying out-of-distribution detection 
of anomalies as a significant deviations in features. Various methods based on this principle were proposed, such as AnoGAN~\cite{Schlegl2017} and its successor f-AnoGAN~\cite{Schlegl2019} that utilize Generative Adversarial Networks, or a deep-metric-learning-based approach with triplet loss that learns features of non-anomalous samples~\cite{Staar2018}, or approach that transfers pre-trained discriminative latent embedding into a smaller network using knowledge transfer for out-of-distribution detection~\cite{uninformedStudents}, termed Uninformed Students. The latter achieved state-of-the-art results in unsupervised anomaly detection on the MVTec dataset~\cite{Bergmann2019}, which, however, only partially reflects the complexity of real-world industrial examples.

\paragraph{Weakly supervised learning}
Various weakly supervised deep-learning approaches have been developed in the context of semantic segmentation and object detection~\cite{Pathak2014,Saleh2016,Bearman2016,Li2018,Wan2018,Wan2019}. In early applications, convolutional neural networks were trained with image-tags using Multiple Instance Learning (MIL)~\cite{Pathak2014} or with constrained optimization as in Constrained CNN~\cite{Pathak2015}. The approach by Seleh et al.~\cite{Saleh2016} further used dense conditional random fields to generate foreground/background masks that act as priors on an object, while Bearman et al.~\cite{Bearman2016} used a single-pixel point label of object location instead of image-tags. Ge et al.~\cite{Ge2020} used a segmentation-aggregation framework learned from weakly annotated visual data and applied it to insulator detection on power transmission lines. Others utilized class activation maps (CAM)~\cite{Zhou2015}. Zhu et al.~\cite{Zhu2019} applied CAM for instance segmentation, while Diba et al.~\cite{Diba2017} simultaneously addressed image classification, object detection, and semantic segmentation, where CAM from image classification is used in a separate cascaded network to improve the last two tasks. 

Class activation maps were also applied to anomaly detection. Lin et al.~\cite{Lin2018} addressed defect detection in LED chips using CAM from the AlexNet architecture~\cite{Zhou2015} to localize the defects, but learning only on the image-level labels. Zhang et al.~\cite{CADN} extended CAM for defect localization with bounding box prediction in their proposed CADN model. Their model directly predicts the bounding boxes from category-aware heatmaps while also using knowledge distillation to reduce the complexity of the final inference model. However, both methods do not consider pixel-level labels in the learning process, thus failing to utilize this information when available.

\paragraph{Mixed supervision} Several related approaches also considered learning with different precision of labels. Souly et al.~\cite{Souly2017} combined fully labeled segmentation masks with unlabeled images for pixel-wise semantic segmentation tasks. They train the model in adversarial manner by generating images with GAN and include any provided weak, image-level labels to the discriminator in GAN that further improves the semantic segmentation. Mlynarski et al.~\cite{Mlynarski2019} addressed the problem of segmenting brain tumors from magnetic resonance images. They proposed to use fully segmented images and combine them with weakly annotated image-level information. 
They focus on the goal of segmenting brain tumor images, while our primary concern is image-level anomaly detection in the industrial surface-defect-detection domain. They also do not perform any analysis of different mixtures of the weakly and fully supervised learning, which is the central point of this paper.

\begin{figure}
    \centering
    \includegraphics[width=\columnwidth]{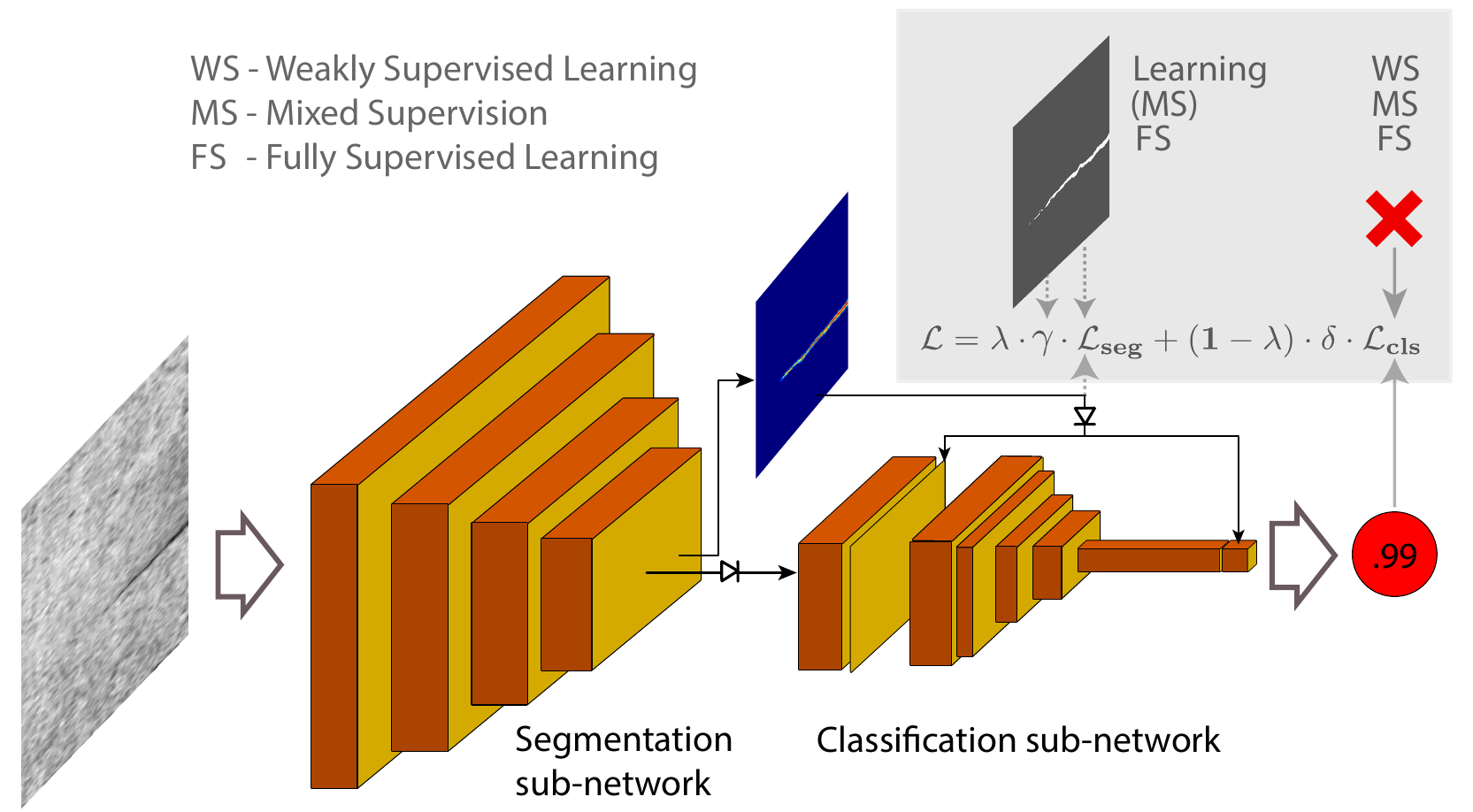}
    \caption{Proposed architecture with two sub-networks suitable for mixed supervision.} 
  \label{fig:arch}
\end{figure}

\section{Anomaly detection with mixed supervision} \label{sec:method}

In this section, we present a deep-learning model that can be trained on a mixture of fully (pixel-level) and weakly (image-level) labeled data.
Learning on weakly labeled data requires a model capable of utilizing segmentation ground-truth/masks when they are available, however, it also needs to utilize image-level, i.e. class-only, labels as well.
The proposed model is based on our previous architecture with two sub-networks~\cite{Tabernik2019JIM, bozic2020}, in which the segmentation sub-network utilizes fine pixel-level information and the classification sub-network utilizes coarse image-level information.
The overall architecture with two sub-networks is depicted in Fig.~\ref{fig:arch}, with the architectural details of each sub-network shown in Tab.~\ref{tab:arch}.

\subsection{Mixed supervision with end-to-end learning}

Learning from a mixture of weakly labeled and fully labeled samples is only possible when both the classification and the segmentation sub-networks are trained simultaneously.

\begin{table}[]
\footnotesize
    \resizebox{\columnwidth}{!}{%
    \centering
            \begin{tabular}{lcc|lcc}
        \toprule
        \multicolumn{3}{c}{\textit{\textbf{Segmentation sub-network}}}		& \multicolumn{3}{c}{\textit{\textbf{Classification sub-network}}}	\\
        \midrule
    	\textit{Layer}	        & \textit{Kernel size}  & \textit{Features}	& \textit{Layer}	            & \textit{Kernel size}	&	\textit{Features}	\\
    	\midrule
    	\multicolumn{2}{l}{\textit{Input:} image} & 3/1 & \multicolumn{2}{l}{\textit{Input:} $[S_f, S_h$]} & 1025	\\
    	\midrule
    	2x Conv2D	    & 5x5	& 32	& Max-pool    	    & 2x2	& 1025	\\
    	Max-pool    	& 2x2	& 32	& Conv2D	        & 5x5	& 8	\\
    	\midrule
    	3x Conv2D   	& 5x5	& 64	& Max-pool    	    & 2x2	& 8	\\
    	Max-pool    	& 2x2	& 64	& Conv2D	        & 5x5	& 16	\\
    	\midrule
    	4x Conv2D   	& 5x5	& 64	& Max-pool	        & 2x2	& 16	\\
    	Max-pool    	& 2x2	& 64	& Conv2D ($C_f$)    & 5x5	& 32	\\
    	\midrule
        Conv2D ($S_f$)  &	5x5	&	1024 & 
        \multicolumn{2}{l}{\makecell[l]{$\left[\mathcal{G}_a(C_f),\mathcal{G}_m(C_f),\mathcal{G}_a(S_h),\mathcal{G}_m(S_h)\right]$}} &	66	\\
        Conv2D ($S_h)$	&	1x1	&	1	& \multicolumn{2}{l}{Fully connected ($C_p$)}    & 1	\\
    	\bottomrule
        \end{tabular}
    }
    \caption{Architecture details for segmentation and classification sub-networks, in which \textit{Features} column represent the number of output features.     $\mathcal{G}_{a}$ and $\mathcal{G}_{m}$ represent global average and max pooling operations. Outputs of the network are a segmentation map $S_h$ and a classification prediction $C_p$.
    }
    \label{tab:arch}
\end{table}
\paragraph{End-to-end learning}
We combine the segmentation and the classification losses into a single unified loss, which allows for a simultaneous learning in an end-to-end manner. The combined loss is defined as:
\begin{equation} \label{eq:loss_total}
    \mathcal{L}_{total} = \lambda \cdot \gamma \cdot \mathcal{L}_{seg} + (1-\lambda) \cdot \delta \cdot  \mathcal{L}_{cls},
\end{equation}
where $\mathcal{L}_{seg}$ and $\mathcal{L}_{cls}$ represent segmentation and classification losses, respectively. For both, we use the cross-entropy loss. The remaining parameters are: $\delta$ as an additional classification loss weight, $\gamma$ as an indicator of the presence of pixel-level annotation and $\lambda$ as a balancing factor that balances the contribution of each sub-network in the final loss. Note that $\lambda$, $\gamma$ and $\delta$ do not replace the learning rate $\eta$ in SGD, but complement it.
$\delta$ enables us to balance the contributions of both losses, which can be on different scales since the segmentation loss is averaged over all pixels, most of which are non-anomalous.

We also address the problem of learning the classification network on the initial unstable segmentation features by learning only the segmentation at the beginning and gradually progress towards learning only the classification at the end. We formulate this by computing their balancing factors as a simple linear function:
\begin{align}
    \lambda &= 1 - \nicefrac{n}{n_{ep}},
\end{align}
where $n$ is the current-epoch index and $n_{ep}$ is the number of all training epochs. Without the gradual balancing of both losses, the learning would in some cases result in exploding gradients. We term the process of gradually shifting the training focus from segmentation to classification as the \textit{dynamically balanced loss}. 
Additionally, using lower $\delta$ values can further reduce the issue of learning on the noisy segmentation features early on.

\paragraph{Using weakly labeled data}
The proposed end-to-end learning pipeline is designed to enable utilisation of the weakly labeled data alongside fully labeled one. Such adoption of mixed supervision allows us to take a full advantage of any pixel-level labels when available, which weakly and unsupervised methods are not capable of using. We use $\gamma$ from Eq.~\ref{eq:loss_total} to control the learning of the segmentation sub-network based on the presence of pixel-level annotation:
\begin{equation}
    \gamma = \begin{cases} 
    1 &\text{ negative image, }\\
    1 &\text{ pos. image with pixel-level label,  }\\
    0 &\text{ pos. image with no pixel-level label.  }\\
    \end{cases}
\end{equation}
We disable the segmentation learning only when the pixel-level label for an anomalous (positive) image is not available. For non-anomalous (negative) training samples, the segmentation output should always be zero for all pixels, therefore, segmentation learning can still be performed.
This allows us to treat non-anomalous training samples as fully labeled samples and enables us to train the segmentation sub-network in the supervised  mode even in the absence of the pixel-level annotations during weakly supervised learning in the case of defect-free images.

\paragraph{Gradient-flow adjustments}

We stop the gradient-flow from the classification layers through the segmentation sub-network, which is required to stabilize the learning and enables training in a single end-to-end manner. During the initial phases of the training, the segmentation sub-network does not yet produce meaningful outputs, hence, neither does the classification sub-network, therefore, error gradients back-propagating from the classification layers can negatively affect the segmentation part. We propose to completely stop those gradients, thus preventing the classification sub-network from changing the segmentation layers. We achieve this by stopping the error back-propagation at two points that connect segmentation and classification networks. The primary point is the use of segmentation features in the classification network. This is depicted by bottom diode symbol in Fig.~\ref{fig:arch}. The second point, at which we stop error gradients is the max/avg-pooling shortcut used by the classification sub-network. This is depicted with the top diode symbol in Fig~\ref{fig:arch}. Those shortcuts utilize the segmentation sub-network's output map to speed-up the classification learning. Propagating gradients back through them would add error gradients to the segmentation's output map, however, this is unnecessary due to the already available pixel-level ground-truth for those features. Without gradient-flow adjustments, we observed a drop in performance in fully supervised scenario and greater instabilities during the training in weak and mixed supervision.

\begin{figure}
    \begin{center}
        \includegraphics[width=\columnwidth]{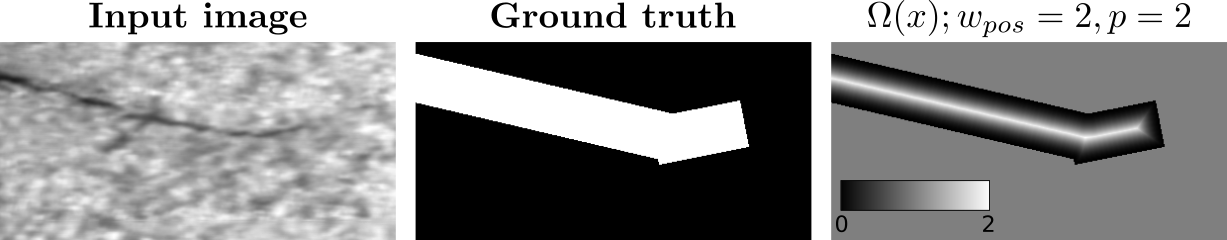}
    \end{center}
    \caption{Segmentation loss weight mask obtained by applying distance transform algorithm on the label. Whiter shades on the segmentation loss mask indicate pixels with greater weight.} 
  \label{fig:seg_loss_mask}
\end{figure}
\subsection{Considering spatial label uncertainty}

When only approximate, region-based labels are available, such as shown in Fig.~\ref{fig:seg_loss_mask}, we propose to consider different pixels of the annotated defective regions differently. In particular, more attention is given to the center of the annotated regions and less to the outer parts. This alleviates the ambiguity arising at the edges of the defects where it is very uncertain whether the defect is present or not. This is implemented by weighting the segmentation loss accordingly. We weight the influence of each pixel at positive labels in accordance with its distance to the nearest negatively labelled pixel by using the distance transform algorithm. 

We formulate weighting of the positive pixels as:
\begin{equation}\label{eq:EDT}
    \mathcal{L}_{seg}(pix) = \Omega\left(\frac{\mathcal{D}(pix)}{\mathcal{D}(pix_{max})}\right) \cdot \mathcal{\hat L}(pix),
\end{equation}
where $\mathcal{\hat L}(pix)$ is the original loss of the pixel, $\nicefrac{\mathcal{D}(pix)}{\mathcal{D}(pix_{max})}$ is the distance to the nearest negative pixel normalized by the maximum distance value within the ground-truth region and $\Omega(x)$ is a scaling function that converts the relative distance value into the weight for the loss. In general, the scaling function $\Omega(x)$ can be defined differently depending on the defect and annotation type, however, we have found that a simple power function provides enough flexibility for different defect types:
\begin{equation}
    \Omega(x) = w_{pos} \cdot x^p,
\end{equation}
where $p$ controls the rate of decreasing the pixel importance as it gets further away from the center, while $w_{pos}$ is an additional scalar weight for all positive pixels. We have often found $p=1$ or $p=2$ as best performing, depending on the annotation type. Examples of a segmentation mask and weights are depicted in Fig.~\ref{fig:seg_loss_mask}. Note, that weights for negatively labeled pixels remain $1$.

\begin{table}[]
\resizebox{\columnwidth}{!}{%
    \centering
\begin{tabular}{llcccc}
        \toprule
        \textit{Dataset}            & \textit{Subset}  & \textit{\# images}  & \textit{\makecell{\# classes}} & \textit{\makecell{\# defect\\types}}  & \textit{Anotations}  \\
            \midrule
        \multirow{2}{*}{\textit{\textbf{\makecell{DAGM\\1-6}}}}    & \textit{positive}     & 450  & \multirow{2}{*}{6} & \multirow{2}{*}{6} & \multirow{2}{*}{ellipse} \\
                                    & \textit{negative} & 3000   &  & &  \\
        \vspace{-1em}\\
        \multirow{2}{*}{\textit{\textbf{\makecell{DAGM\\7-10}}}}    & \textit{positive}     & 600 & \multirow{2}{*}{4} & \multirow{2}{*}{4} & \multirow{2}{*}{ellipse} \\
                                    & \textit{negative} & 4000   &  & &  \\
        \vspace{-0.5em}\\
        \multirow{2}{*}{\textit{\textbf{KSDD}}}    & \textit{positive}     & 52 & \multirow{2}{*}{1} & \multirow{2}{*}{1} & \multirow{2}{*}{\makecell{rotated\\bounding box}} \\
                                    & \textit{negative} & 347    &  & & \\
        \vspace{-0.5em}\\
        \multirow{2}{*}{\textit{\textbf{KSDD2}}}    & \textit{positive}     & 356 & \multirow{2}{*}{1} & \multirow{2}{*}{$>5$} & \multirow{2}{*}{\makecell{fine}} \\
                                    & \textit{negative} & 2979   &  & &  \\
        \vspace{-0.5em}\\
        \multirow{2}{*}{\textit{\textbf{\makecell[l]{Severstal\\Steel}}}}    & \textit{positive}     & 4759  & \multirow{2}{*}{1} & \multirow{2}{*}{$>5$} & \multirow{2}{*}{\makecell{fine or rotated\\bounding box}} \\
                                    & \textit{negative} & 6666   & & &  \\
        \bottomrule                            
        \end{tabular}
    }
    \caption{Details of the the evaluation datasets.}
    \label{tab:dataset-sizes}
    
\end{table}

\section{Experimental setup} \label{sec:experiment-setup}

In this section, we detail the datasets and the performance metrics used to evaluate the proposed method as well as provide additional implementation details.

\subsection{Datasets} \label{sec:datasets}

\begin{figure}
    \begin{center}
        \includegraphics[width=\columnwidth]{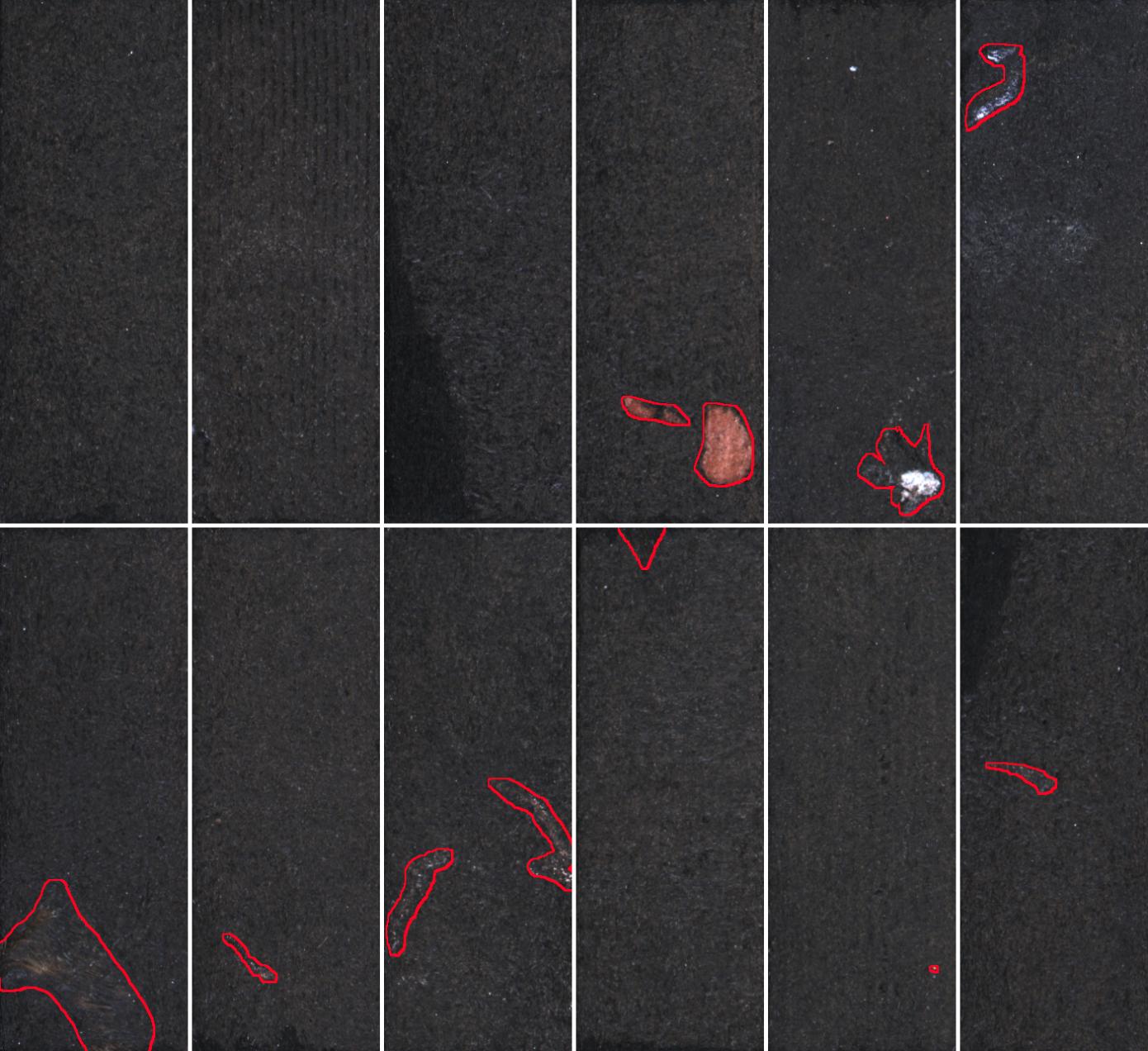}
    \end{center}
    \caption{Examples of training images from KolektorSDD2 
    dataset with pixel-wise annotations shown in red-overlay region.
    } 
    \label{fig:samples_ksdd2_train}
\end{figure}

We performed an extensive evaluation of the proposed method on four benchmark datasets. The summary of individual datasets is shown in Tab.~\ref{tab:dataset-sizes}, while additional details are provided below. A couple of images from all four datasets are presented in Figs.~\ref{fig:samples_ksdd2_train}, \ref{fig:samples_dagm}, \ref{fig:samples_ksdd}, \ref{fig:samples_ksdd2} and \ref{fig:samples_steel}.

\paragraph{\textbf{DAGM}}
The DAGM~\cite{Weimer2016} dataset is a well-known benchmark dataset for surface-defect detection. 
It contains grayscale images of ten different computer-generated surfaces and various defects, such as scratches or spots.
Each surface is treated as a binary-classification problem. Initially, six classes were presented while four additional ones were introduced later; consequently, some related methods report results only on the first six classes, while others report on all ten of them. 

\paragraph{\textbf{KolektorSDD}}
The KolektorSDD~\cite{Tabernik2019JIM} dataset contains grayscale images of a real-world production item; many of them contain visible surface cracks. Due to the small sample size, the images are split into three folds as in~\cite{Tabernik2019JIM}, while final results are reported as an average of three-fold cross validation.

\paragraph{\textbf{KolektorSDD2}}
Since the above mentioned datasets have been practically solved, and there is a huge need for real-world, reliable, complex and well annotated surface-detection datasets that would enable a fair comparison between different approaches, we compiled a novel dataset Kolektor Surface-Defect Dataset 2, abbreviated as KolektorSDD2\footnote{The dataset is publicly available at \url{https://www.vicos.si/Downloads/KolektorSDD2}}.
The dataset is constructed from color images of defective production items, captured with a visual inspection system, that were provided and partially annotated by our industrial partner Kolektor Group d.o.o.
Images for the proposed dataset were captured in a controlled environment and are of similar size, approximately 230 pixels wide and 630 pixels high. 
The dataset is split into the train and the test subsets, with 2085 negative and 246 positive samples in the train, and 894 negative and 110 positive samples in the test subset.
Defects are annotated with fine-grained segmentation masks and vary in shape, size and color, ranging from small scratches and minor spots to large surface imperfections.
Several images from this dataset are shown in Fig.~\ref{fig:samples_ksdd2_train}.

\paragraph{\textbf{Severstal Steel defect dataset}}

The Severstal Steel defect dataset~\cite{SeverstalSteel2019} is significantly larger than the other three datasets, containing 12,568 grayscale images in 4 classes, with various kinds of defects. We use only a subset of the dataset in our evaluation. In particular, we use all negative images but consider only positive images with the most common defect class present in the image (class 3). 
The defects are very diverse in size, shape and appearance, ranging from scratches and dents to excess material.
Although the dataset is fairly large and diverse, some defects are quite ambiguous and may not be properly annotated as observed in the third example in Fig.~\ref{fig:samples_steel}.

\subsection{Performance metrics}

In all experiments, we focus on evaluating per-image classification metrics, which are the most relevant metrics in the industrial quality control, since they decide whether to keep or discard the inspected items. In particular, we mostly measure the performance in terms of the average precision (AP), which is calculated as the area under the precision-recall curve. For the DAGM dataset, we also report several other metrics that are used to report the results in related literature, to enable a fair comparison of the proposed method with related approaches.

\subsection{Implementation details}

The proposed architecture is implemented\footnote{Implementation is available on github: \url{https://github.com/vicoslab/mixed-segdec-net-comind2021}} in PyTorch framework. In all experiments, the network is trained with stochastic gradient descent, with no momentum and with no weight decay. 
For all experiments, we disable the gradient flow from classification network to segmentation network and employ loss weighting for positive pixels.
We enable dynamically balanced loss for all cases with mixed supervision and fully supervised learning, but disable it when only weak labels are used. Since pixel-level labels are never available in weakly supervised case, it has proven more useful to immediately start learning classification layers instead of delaying it with the gradual increase of $\lambda$.
To account for the unbalanced datasets, we employ undersampling of negative samples, in each training epoch we sample a selection of negative samples of equal size as the positive subset. 
We also ensure that over the course of learning, all negative images are used approximately as often.
We also apply morphological dilation to the annotations in datasets where they are narrow and small in order to enlarge the label and facilitate distance-transform weighting. This has often proven better than using narrow annotations, as small defects can be difficult to precisely annotate, and, for some types of defects, the boundary between the defect and the background can not be precisely defined. Furthermore, distance transform assigns very small weights to the pixels near the edge of defects, thus dilating the annotations increases the weights for those pixels.

\subsection{Training and inference time}
The proposed method is capable of real-time detection, achieving 57 fps on $512\times512$ images (DAGM), 23 fps on $512\times1408$ images (KSDD), 85 fps on $232\times640$ images (KSDD2) and 36 fps on $1600\times256$ images (Severstal Steel), running on a single Nvidia RTX2080Ti GPU. The training time depends on the dataset size. In our experiments, training a model on DAGM took 35 minutes (7 or 10 classes), on Severstal Steel ($N_{all}=3000$) it took 460 minutes, while for KSDD2 and KSDD (for a single fold) it took 15 and 32 minutes respectively.

\subsection{Related methods}

We compare our approach against several related methods that have reported their results on DAGM~\cite{Staar2018, CADN, Kim2017_DAGM, Racki2018, Weimer2016, Lin2020, Wang2018c, Huang2020, Liu2020a} and KolektorSDD~\cite{Dong2020, Tabernik2019JIM, Liu2020} in the literature. Additionally, we evaluate two state-of-the-art unsupervised methods that we apply to all four datasets. Below, we provide more details on the implementation of the two unsupervised methods.

\paragraph{Uninformed students}
Since the code for Uninformed-students~\cite{uninformedStudents} is not publicly available, we implemented our own version, which attains results comparable to the ones reported in the literature. We additionally opted to use a per-image defect score for classification similar to~\cite{Nguyen2019Scoring}, which slightly improved results compared to the original implementation. For each image, we performed a $21\times21$ pixel average pooling with $\text{stride}=1$, to obtain patch scores, and then calculated the final image score as $s=\nicefrac{max(patch)}{(1+avg(pixel))}$, i.e. we took maximum patch score divided by average pixel score plus one.
This increased the AP in all four datasets.

\begin{table}[]
\resizebox{\columnwidth}{!}{%
\begin{tabular}{@{}lccccccccc@{}}
\toprule
\multirow{2}{*}{Method}                & \multirow{2}{*}{Type} & \multicolumn{4}{c}{10 classes} & & \multicolumn{2}{c}{6 classes} \\ \cmidrule(l){3-6} \cmidrule(l){7-9} 
                                       &                       & AP     & AUC   & F1    & CA    & & CA      & mAcc    \\ \midrule
f-AnoGAN~\cite{fAnoGAN}                & US                    & 19.5   & 57.5  & 27.8     & 79.7    & & 81.7     & 54.6    \\
Uninf. stud.~\cite{uninformedStudents} & US                    & 66.8   & 86.4  & 67.1  & 84.3  & & 79.3     & 78.5    \\ 
Staar~\cite{Staar2018}                 & US                    & -      & 83.0  & -     & -     &  & -        & -       \\ \midrule
CADN-W18~\cite{CADN}                               & WS                    & -      & -     & 63.2  & 86.2  & & -        & -       \\
CADN-W18(KD)                           & WS                    & -      & -     & 65.8  & 87.6  & & -        & -       \\
CADN-W32~\cite{CADN}                   & WS                    & -      & -     & 69.0  & 89.1  & & -        & -       \\
\rowcolor{Gray}
Ours (N=0)                          & WS                    & 74.0   & 86.1  & 74.6  & 89.7  & & 85.4    & 81.4    \\ \midrule
\rowcolor{Gray}
Ours (N=5)                          & MS                    & 91.5   & 94.9  & 92.3  & 92.9  & & 88.1    & 91.6    \\
\rowcolor{Gray}
Ours (N=15)                         & MS                    & 100    & 100   & 100   & 100   & & 100      & 100     \\
\rowcolor{Gray}
Ours (N=45)                         & MS                    & 100    & 100   & 100   & 100   & & 100      & 100     \\ \midrule
\rowcolor{Gray}
Ours (N=$N_{all}$)                        & FS                    & 100    & 100   & 100   & 100   &     & 100      & 100     \\
Kim~\cite{Kim2017_DAGM}                & FS                    &  -     & -     & -     & -     &  & -      & 99.9    \\
Rački~\cite{Racki2018}                 & FS                    &  -     & 99.6  & -     & 99.6     & & 99.2    & 99.4    \\
Weimer~\cite{Weimer2016}               & FS                    &  -     & -     & -     & -     & & -     & 99.2    \\
Lin~\cite{Lin2020}                     & FS                    &  -     & 99.0  & -     & -     & & -      & 99.8    \\
Wang~\cite{Wang2018c}                  & FS                    &  -     & -     & -     & -     & & 99.8     & 99.4    \\
Huang~\cite{Huang2020}                 & FS                    &  -     & -     & -     & -     &  & -        & 99.8    \\ 
Liu~\cite{Liu2020a} ($N_{all}$=20\%)                  & FS                    &  -     & -     & -     & -     & & 99.9 & -           \\

\bottomrule
\end{tabular}%
}
\caption{Comparison with related work on the DAGM dataset. For AP, AUC, F1-measure and classification accuracy (CA), the results are averaged over all 10 classes, whereas for (second) CA and mAcc=(TPR+TNR)/2, they are averaged only over the first 6 classes for comparison with the related work that report results only in  terms of those metrics. 
}
\label{tab:DAGMrelatedWork}
\end{table}
\paragraph{f-AnoGAN}
We used publicly the available TensorFlow implementation of f-AnoGAN~\cite{fAnoGAN}. In all experiments, we extracted 500 patches per image for training, with each patch $64\times64$ pixels in size. In inference, a patch size of $64\times64$ pixels was extracted and classified for every pixel location, which resulted in a heatmap of the same size as the original image. For scoring, we used the distance between the reconstructed image and the original image as well as between the discriminator features of the reconstructed and the original image. A mean squared error was used to measure both distances. We also opted to use  normalization of the scoring similar to~\cite{Nguyen2019Scoring}, which slightly improved results on noisy outputs. 

\section{Evaluation results} \label{sec:results}

In this section, we present the results of the extensive evaluation of the proposed method on several industrial quality-control problems. We first present the evaluation on the DAGM dataset for comparison to different related works, and then present results on KolektorSDD, KolektorSDD2 and Severstal steel, all of which represent practical surface-anomaly detection problems taken from real-world industrial cases. Finally, we present a detailed ablation study that analyses the performance of individual components in the proposed method.

We simulate and evaluate different levels of supervision by varying the number of positive (i.e., anomalous) training samples for which we have available pixel-wise segmentation masks (i.e, the number of segmented anomalous images $N$ used):

\begin{enumerate}[label=\roman*)]
    \item \textbf{\textit{weak supervision}}, with only image-level labels for all images (i.e., $N=0$),
    \item \textbf{\textit{mixed supervision}}, with image-level labels for all images but also pixel-level labels only for a subset of anomalous images (i.e., $0<N<N_{all}$), and
    \item \textbf{\textit{full supervision}}, with image-level and pixel-level labels for all anomalous images (i.e., $N = N_{all}$).
\end{enumerate}
Although we limit the number of images with the segmentation mask, we always use data with the image-level label, i.e. weak label that only indicates whether the anomaly is present in the sample or not.

\subsection{DAGM}

We first performed evaluation on the DAGM dataset. We consider the number of positive segmented samples $N = \{0, 5, 15, 45, N_{all}\}$, and use only image-level labels for the remaining training images. In all cases, we trained for $n_{ep}=70$ epochs, with the learning rate $\eta=0.05$, batch size $bs=1$, $\delta=1$, $w_{pos}=10$ and $p=1$. We dilated segmentation masks with $7\times7$ kernel. 

\begin{figure}
    \begin{center}
        \includegraphics[width=\columnwidth]{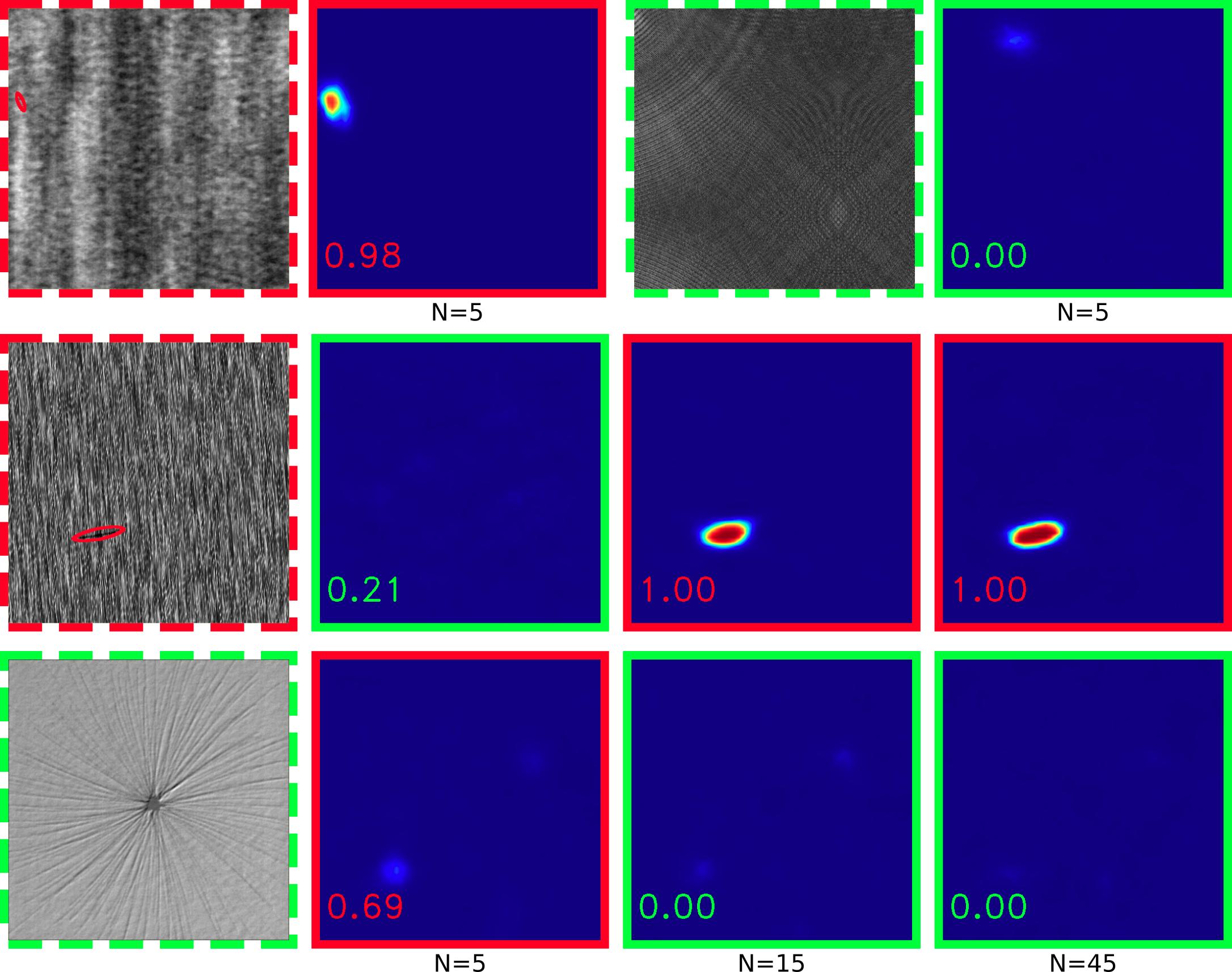}
    \end{center}
    \caption{Examples of segmentation outputs and scores from the proposed method on the DAGM dataset. Number of fully labeled samples is marked with $N$. Border colour indicates presence (red) or absence (green) of defects, with dashed border around ground-truth images and solid border around predictions. The detection score that represents the probability of the defect is indicated in the bottom left corner. 
    } 
    \label{fig:samples_dagm}
\end{figure}

\begin{figure}
    \centering
    \includegraphics[width=\columnwidth]{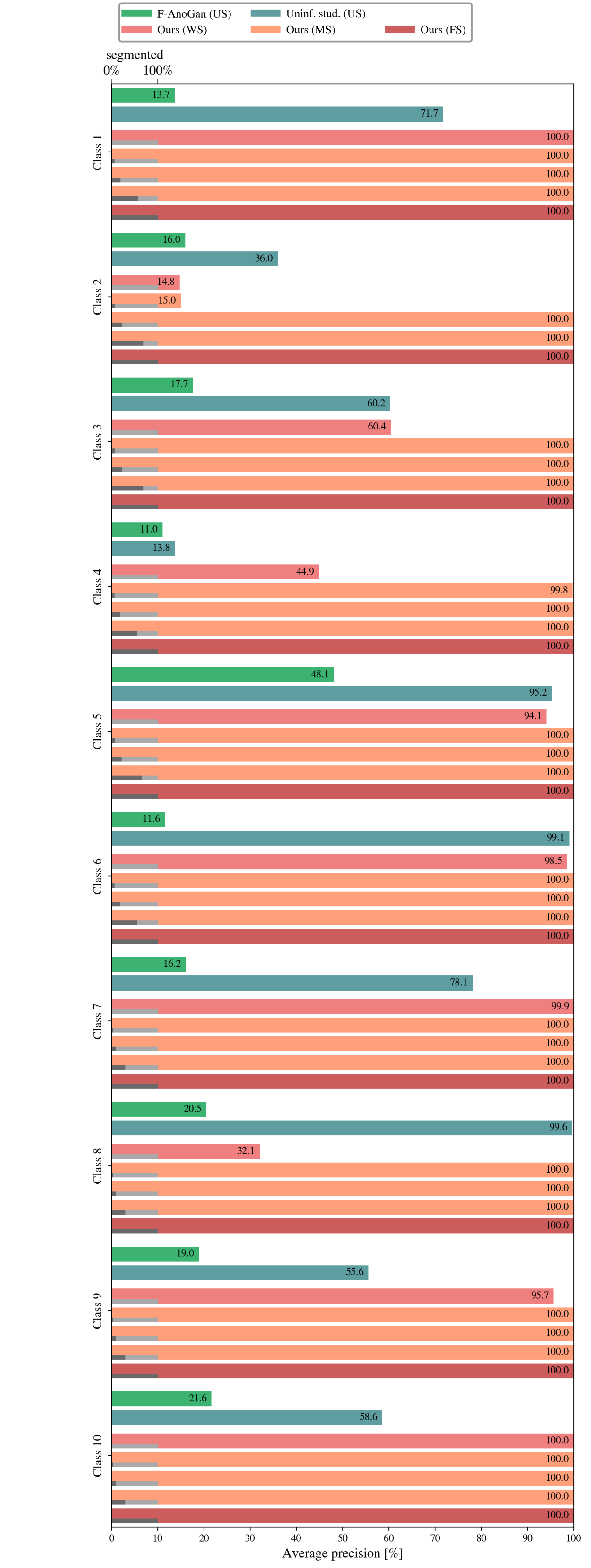}
    \caption{Results on the DAGM dataset in terms of average precision (AP).} 
    \label{fig:dagm1}
\end{figure}

In Fig.~\ref{fig:dagm1}, we show  AP for all 10 classes of the dataset. 
For each class, we show AP of both unsupervised methods and of our model for all different values of $N$.
On the bottom left part of each bar, we also indicate the percentage of the positive samples for which we used pixel-level labels, $\nicefrac{N}{N_{all}}$, which are encoded as proportions of dark and light gray bars.

The proposed method achieves AP of over $90\%$ for 6 classes even with $N=0$ and, on average, achieves $7.2$ percentage points higher AP than the current state-of-the-art unsupervised approach.
Introducing only five pixel-level labels allows the method to achieve $100\%$ detection rate on 8 classes
and raises the average AP to $91.5\%$.
Finally, the method achieves $100\%$ detection rate on all 10 classes with only 15 pixel-level labels, thus outperforming other fully supervised approaches while having less than a quarter of positive samples with pixel-level labels. Several examples of detection are depicted in Fig.~\ref{fig:samples_dagm}.

In Tab.~\ref{tab:DAGMrelatedWork}, we compare the proposed method with a number of other approaches. We report CA and $mAcc=\nicefrac{TPR+TNR}{2}$ metrics averaged over the first six classes since the related methods that use those metrics report results only on the first six classes. Other metrics are averaged over all 10 classes. The proposed method outperforms all other weakly supervised and unsupervised approaches shown at the top of the table and also all fully supervised methods shown at the bottom of the table. Moreover, 100\% detection rate can be achieved in mixed supervision with only 15 fully annotated pixel-level samples. This points to superior performance and flexibility when allowing for mixed supervision.

\begin{figure}
    \begin{center}
        \includegraphics[width=\columnwidth]{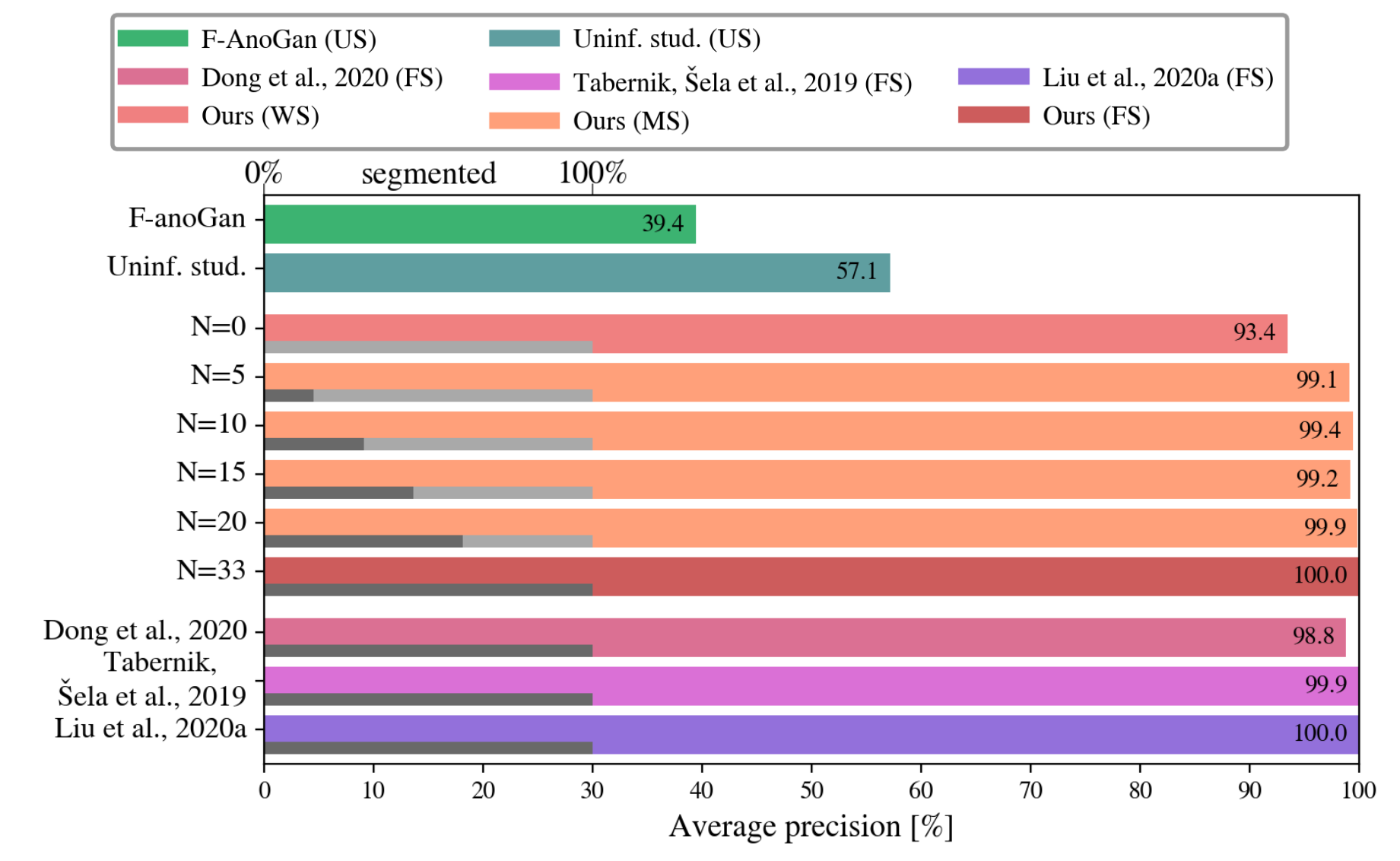}
    \end{center}
    \caption{Results on the KolektorSDD dataset in terms of average precision (AP).}
    \label{fig:ksdd}
\end{figure}

\subsection{KolektorSDD}
On KolektorSDD dataset, we varied the number of segmented positive samples $N = \{0, 5, 10, 15, 20, N_{all}\}$, where $N_{all}=33$ and report the performance in terms of the average precision (AP). Identical hyper-parameters were used for all mixed and fully supervised runs, with $n_{ep}=50$, $\eta=1$, $bs=1$, $\delta=0.01$, $w_{pos}=1$ and $p=2$, whereas we increased $\delta$ to $1$ and decreased $\eta$ to $0.01$ for run with no pixel-level annotations. We dilated segmentation masks with $7\times7$ kernel.  

\begin{figure}
\centering

\begin{subfigure}[b]{\columnwidth}
   \includegraphics[width=1\linewidth]{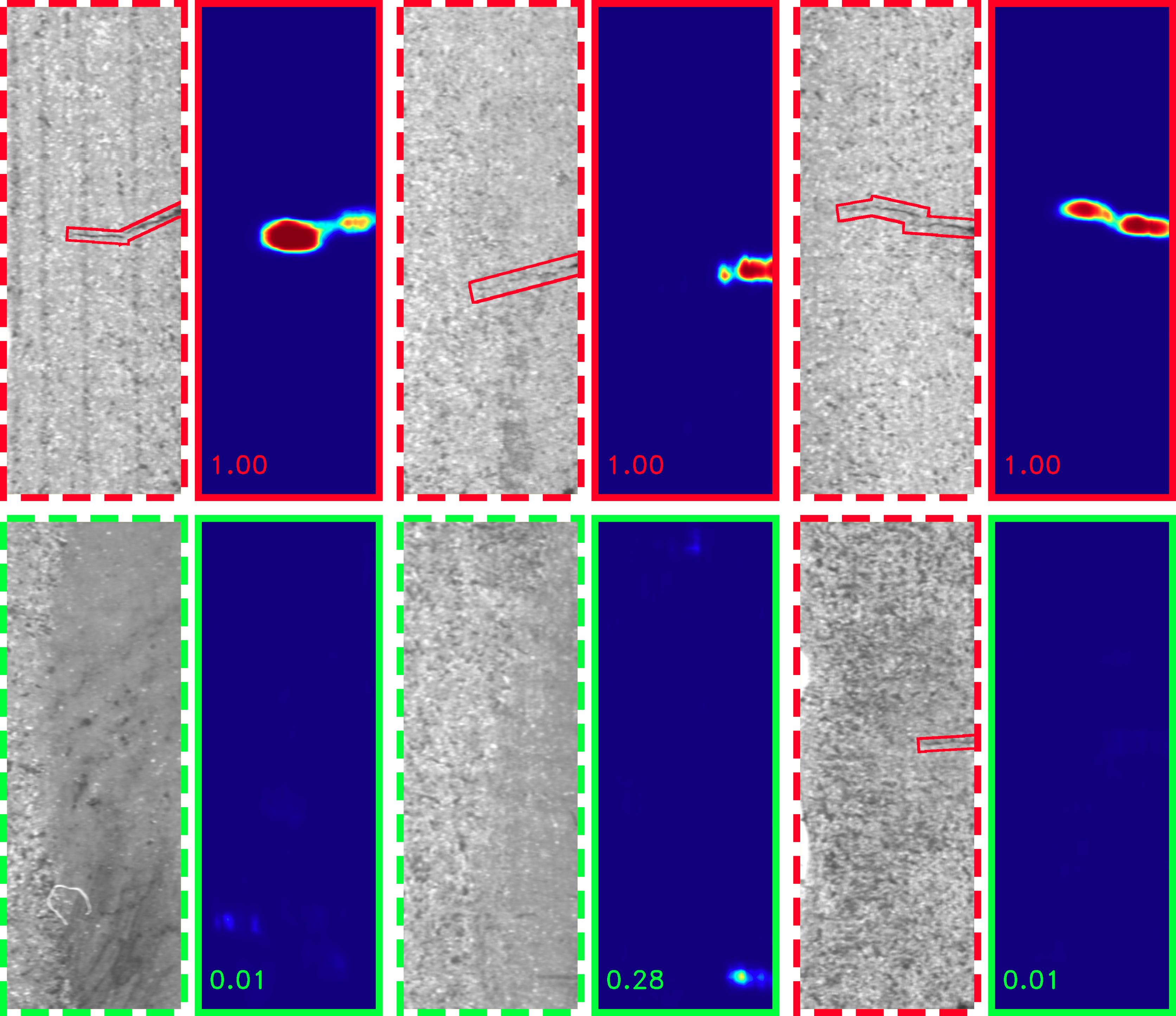}
   \caption{KolektorSDD, $N=5$}
   \label{fig:samples_ksdd}
\end{subfigure}

\begin{subfigure}[b]{\columnwidth}
   \includegraphics[width=1\linewidth]{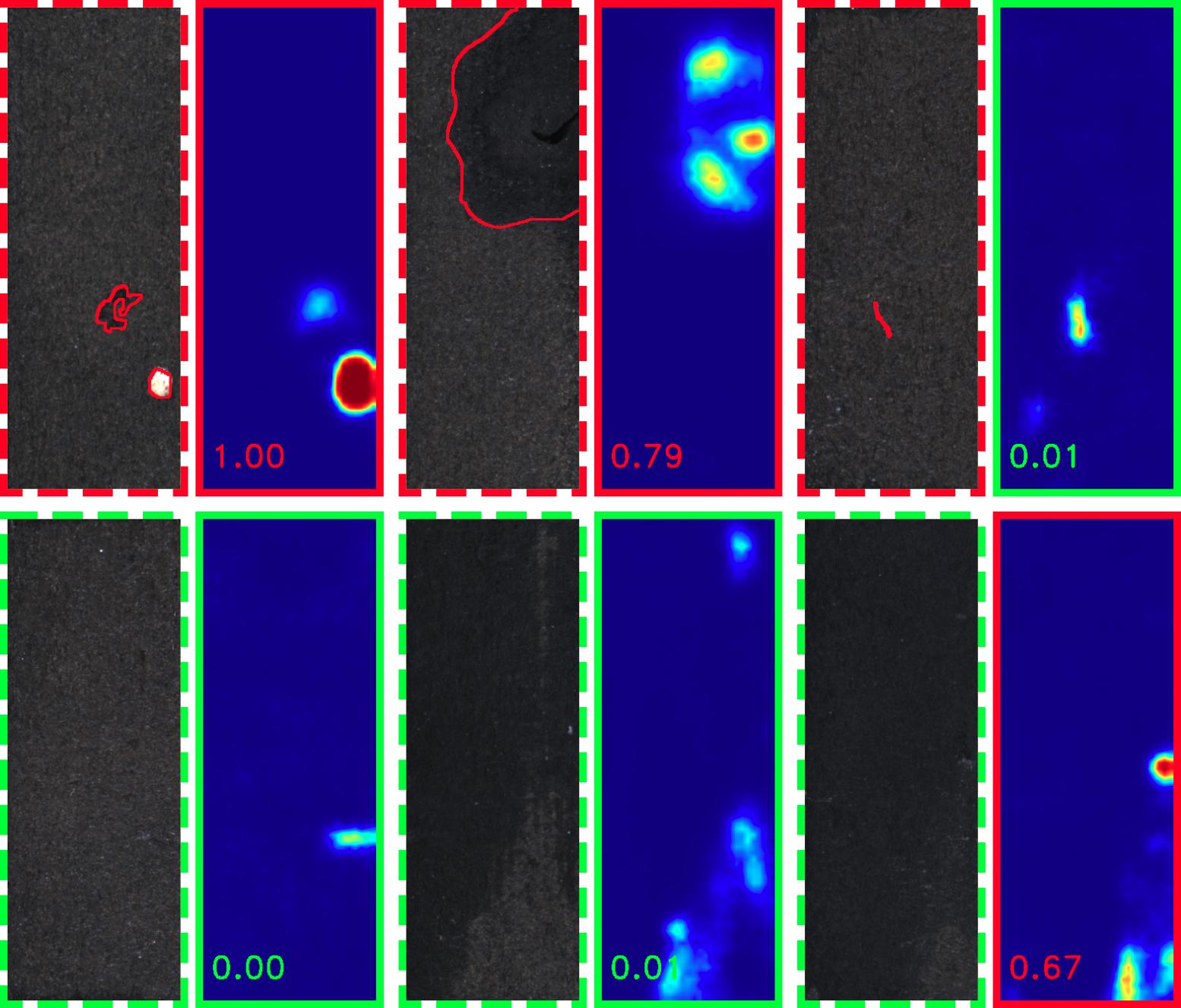}
   \caption{KolektorSDD2, $N=N_{all}$}
   \label{fig:samples_ksdd2}
\end{subfigure}

\begin{subfigure}[b]{\columnwidth}
   \includegraphics[width=1\linewidth]{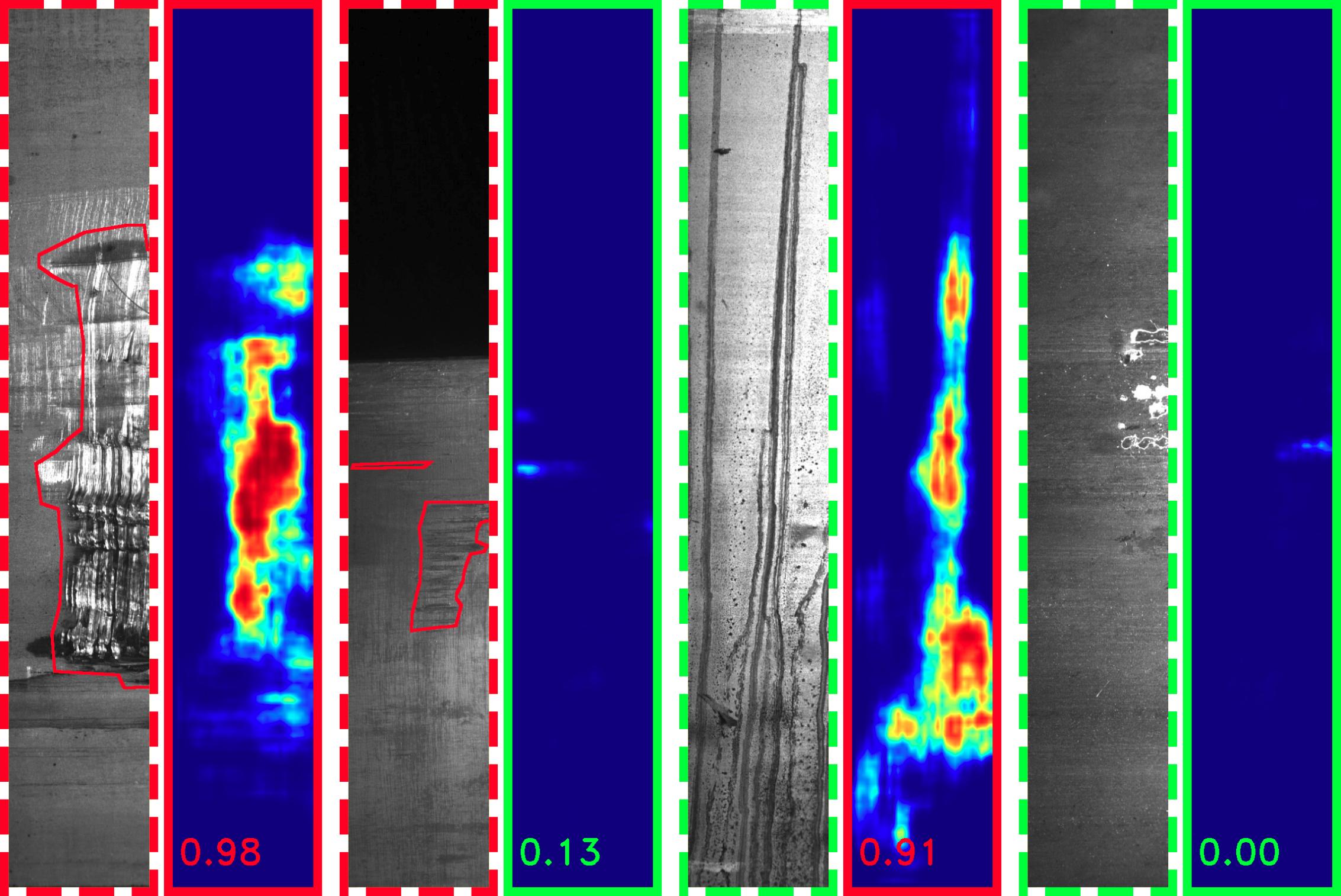}
   \caption{Severstal Steel, $N=N_{all}=1500$}
   \label{fig:samples_steel} 
\end{subfigure}
\caption{Examples of detections from (a) KolektorSDD, (b) KolektorSDD2 and (c) Severstal Steel defect dataset.}
\label{fig:samples_all} 
\end{figure}

In the Fig.~\ref{fig:ksdd}, we show AP for various numbers of segmented positive samples. Without any segmented positive samples, our proposed approach already achieves AP of 93.4\% and significantly outperforms the current state-of-the-art unsupervised methods that do not (and can not) use any pixel-wise information, by roughly doubling their results. 
When only 5 segmented samples are added, the AP drastically increases to 99.1\% and approaches fairly close to that of the fully supervised scenario. This demonstrates well that the full annotation is often not needed for all images and only a few annotated images are enough to achieve respectable results. In this case, less than 15\% of images needed to be annotated, which can significantly reduce the annotation cost with almost no classification loss. When using all fully labeled samples the proposed approach achieves 100\% detection rate.
Several examples of defections are depicted in Fig.~\ref{fig:samples_ksdd}.

\subsection{KolektorSDD2}
Next, we evaluate our approach on the newly proposed KolektorSDD2 dataset. We used identical hyper-parameters for all experiments, $n_{ep}=50$, $\eta=0.01$, $bs=1$, $\delta=1$, $w_{pos}=3$ and $p=2$. Since the segmentation mask are very fine, we used a larger, $15\times15$ kernel for dilation.

\begin{figure}
    \begin{center}
        \includegraphics[width=\columnwidth]{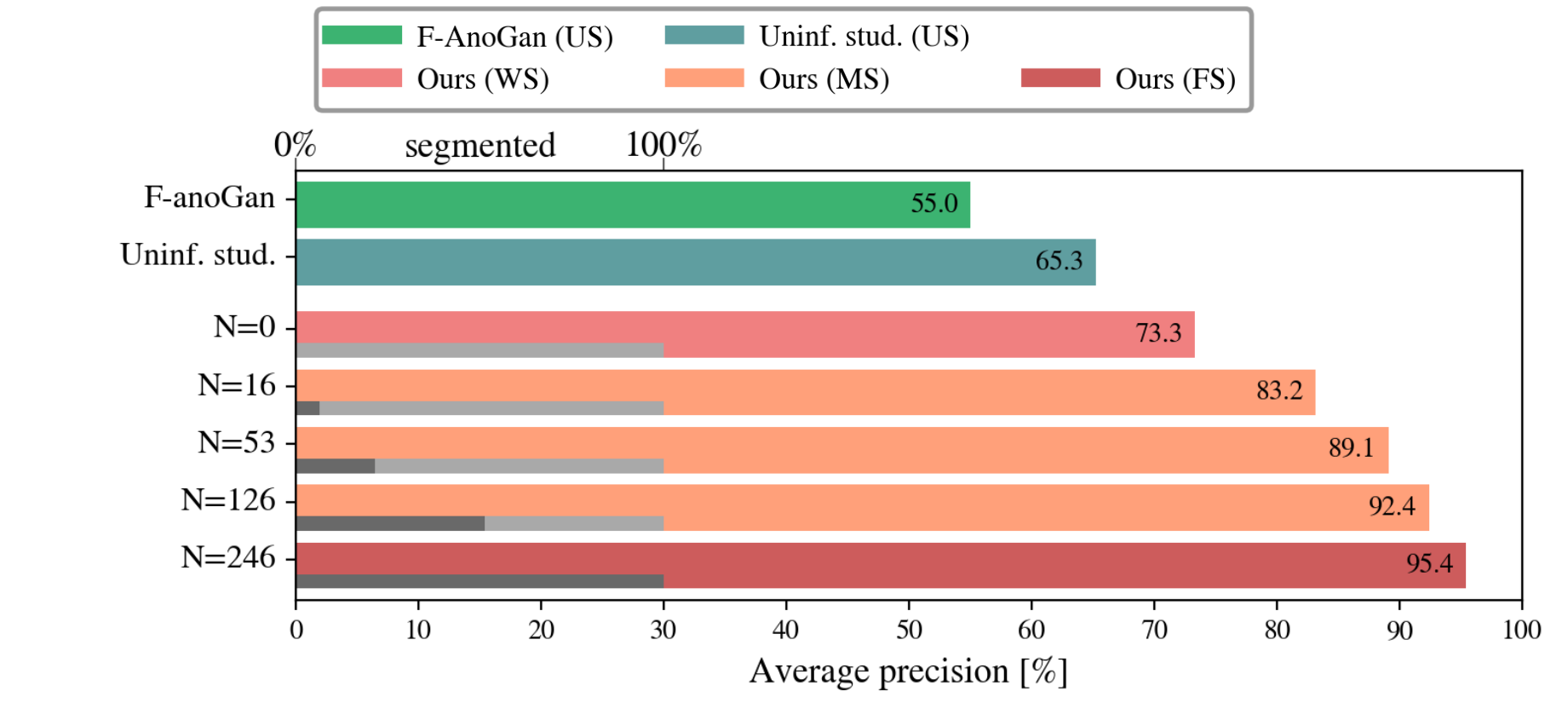}
    \end{center}
    \caption{Results on the KolektorSDD2 dataset in terms of average precision (AP).
    } 
    \label{fig:ksdd2}
\end{figure}

\begin{figure}[h]
    \begin{center}
        \includegraphics[width=\columnwidth]{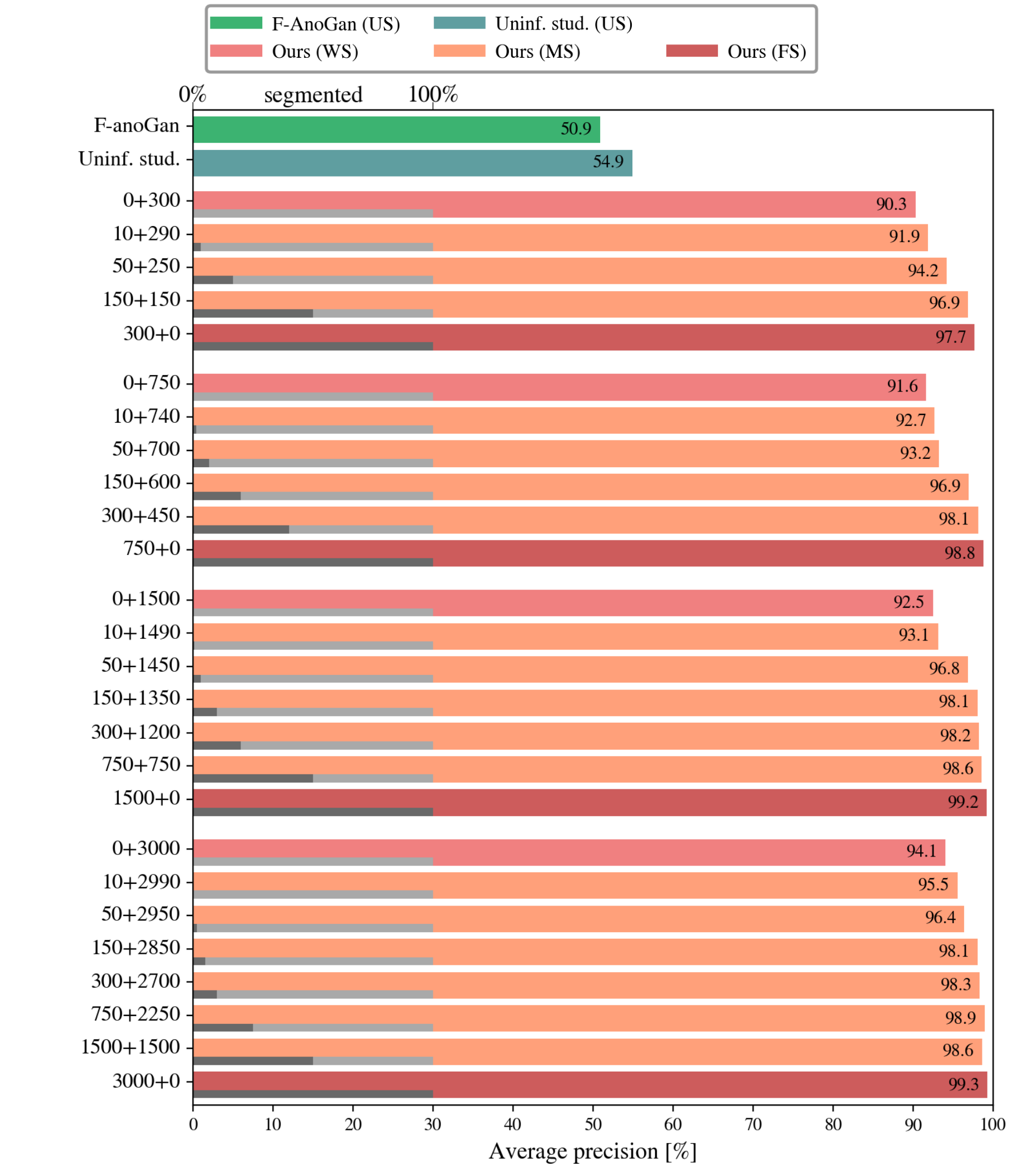}
    \end{center}
    \caption{Results of evaluation on the Severstal steel defect database. The number of $fully +  weakly$ labeled positive samples are shown as labels to the left of the bars.} 
    \label{fig:steel}
\end{figure}

Results obtained on this challenging dataset are presented in Fig.~\ref{fig:ksdd2}, where AP for all values of $N$ are shown.
Even without any pixel-level labels, the proposed approach outperforms current state-of-the-art unsupervised methods and achieves AP of 73.3\%. When 16 segmented samples are introduced, AP increases significantly to 83.2\% and keeps increasing steadily with the introduction of additional pixel-level labels, reaching 95.4\% when $N=N_{all}$. 
This demonstrates that the proposed method scales well with the introduction of segmented samples and is capable of taking the full advantage of all the information about the samples that it is presented with. Several examples of detections are depicted in Fig.~\ref{fig:samples_ksdd2}.

\subsection{Severstal Steel defect dataset}
Lastly, we evaluate the proposed method on Severstal Steel defect dataset. Since this is a large dataset, we also vary the overall number of positive training images available. We used $N_{all} = \{300, 750, 1500, 3000\}$ positive samples, and then for each case used $N = \{0, 10, 50, 150, 300\}$, as well as $N=\{750, 1500, 3000\}$, where $N\leq N_{all}$ that were segmented with pixel-wise accuracy. We used all negative samples regardless of used $N$ and $N_{all}$. We trained for $n_{ep}=\{90, 80, 60, 40\}$ when $N_{all}=\{300, 750, 1500, 3000\}$, respectively, and used $\eta=0.1$, $bs=10$,  $\delta=0.1$, $w_{pos}=1$ and $p=2$.

Fig.~\ref{fig:steel}, shows AP for all $N_{all}$ and $N$.
Bars are grouped according to the $N_{all}$, with the labels showing the number of segmented positive samples and the number of positive samples without pixel-level labels.
We observe that the proposed model is capable of learning from only image-level labeled samples, where it achieves AP of 90.3\%, 91.6\%, 92.5\% and 94.1\% for $N=0$ and $N_{all}=\{300, 750, 1500, 3000\}$.
Note that with the positive set that is 10 times larger, the AP only increases for $3.8$ percentage points (from $90.3$\% to $94.1\%$), pointing to a logarithmic increase in performance with the increase of the number of the positive samples. 
With the introduction of the pixel-level labels, the overall performance increases significantly, particularly for experiments with smaller $N_{all}$.

In the case of $N_{all}=300$, the AP increases from 90.3\% to 94.2\% when only 50 pixel-wise labels were added and surpasses the AP achieved when using 3000 positive samples with image-level labels only. 
Furthermore, we can obtain similar results by using 3000 positive samples with only 150 pixel-level labels as with using 300 positive samples with pixel-level labels. Several examples of detections are depicted in Fig.~\ref{fig:samples_steel}.

\subsection{Ablation study}

Finally, we evaluate the impact of the individual components, namely the \textit{dynamically balanced loss}, the \textit{gradient-flow adjustment} and the \textit{distance transform}, in the proposed model. We perform the ablation study on the DAGM, KolektorSDD, and Severstal Steel Defect datasets, however, for the later we use only a subset of images to reduce the training time, we used 1,000 positive and negative samples for training and the same amount for testing, we trained for 50 epochs.

We report the performance by gradually enabling individual components, and by disabling a specific component while leaving all the remaining ones enabled. Results are reported in Tab.~\ref{tab:ablation}. The results indicate that on all three datasets the worst performance is achieved without any component enabled while the best performance is achieved with all three components used. In the latter case, the proposed model is able to completely solve the DAGM and KolektorSDD datasets while achieving AP of 98.74\% on Severstal Steel dataset. The second part of Tab.~\ref{tab:ablation} also demonstrates the importance of each component for weakly supervised learning and for mixed supervision. In both cases, the best result on all three datasets is achieved only with all components enabled. Note that for weakly supervised learning, we did not have any pixel-level annotations and, therefore, could not apply the distance transform. We describe the contribution of each component to the overall improvements in more details below.

\paragraph{Dynamically balanced loss}

Enabling only dynamic balancing of losses with gradual inclusion of classification network already provides a boost of the performance in all three datasets. In fully supervised learning, the dynamically balanced loss improves AP by 1.08 percentage points in DAGM (from 89.02\% to 90.10\%), by 0.23 points in KolektorSDD (from 99.77\% to 100.00\%) and by 1.68 points in Severstal Steel (95.36\% to 97.04\%). Similar improvements can also be observed in mixed supervision, with 4.44, 0.75 and 1.99 percentage points improvements for DAGM, KolektorSDD and Severstal Steel.

\paragraph{Gradient-flow adjustment}

The gradient-flow adjustment has proven to be equally important as the dynamically balanced loss. Both improvements naturally result in a similar performance in KolektorSDD and Severstal Steel since they both prevent unstable segmentation features to significantly affect learning of the classification layers in the early stages. 
However, enabling both improvements is more robust as it eliminates the convergence issues while also improving results on all three datasets, especially for DAGM, in which 100\% detection rate can be achieved for fully supervised learning and AP of 95.37\% for mixed supervision. On the other hand, it has also proven better to use only gradient-flow adjustment in weak supervision and to completely disable dynamically balanced loss, since in this case segmentation loss is not present.

\paragraph{Spatial label uncertainty}

Lastly, enabling the distance transform as weights for the positive pixels pushes performance on KolektorSDD to 100\% detection rate, therefore completely solving KolektorSDD and DAGM in fully supervised mode. It also improves AP on Severstal Steel dataset from 98.24\% to 98.74\%. Moreover, distance transform is even more important in mixed supervision where it enables 100\% detection rate with only 25\% of fully annotated data and using only weak labels for the remaining data. 

\begin{table*}[t]
\caption{Performance of individual components on three datasets by gradually including each one for fully supervised (FS), mixed supervision (MS), and weakly supervised (WS) learning. We report average precision (AP) and the number of false positives (FP) and false negatives (FN).}
\label{tab:ablation}
\centering
\resizebox{0.9\textwidth}{!}{%
\begin{tabular}{lcccccccccccc}
\toprule
    &\multicolumn{2}{c}{\textit{DAGM}} && \multicolumn{2}{c}{\textit{KolektorSDD}} && \multicolumn{2}{c}{\textit{Severstal Steel}} & \multirow{2}{*}{\textit{\makecell[c]{Dynamically\\balanced loss} } } & \multirow{2}{*}{\textit{\makecell[c]{Gradient-flow\\adjustment}}}&  \multirow{2}{*}{\textit{\makecell[c]{Distance\\transform} } } \\
    \cmidrule{2-3} \cmidrule{5-6} \cmidrule{8-9} \vspace{-8pt}\\
    & AP & FP+FN && AP & FP+FN && AP & FP+FN &  &   & \\ 
\midrule
    \multirow{6}{*}{\rotatebox{90}{{\textit{\makecell{FS\\(N=100\%)}}}}} 
    & 89.02 & 1010+45 && 99.77 & 0+1 && 95.36 & 108+121 &  &  & \\

    & 90.10 & 995+14 && 100.00 & 0+0  && 97.04 & 95+76 & \checkmark &   &  \\ 

    & 100.00 & 0+0 && 99.88 & 1+0 && 98.24 & 52+58 & \checkmark & \checkmark  &  \\ 
    & 91.88 & 471+68 && 99.88 & 1+0 && 97.80 & 57+60 & \checkmark &   & \checkmark \\ 
    & 99.95 & 2+2 && 99.75 & 1+0 && 98.32 & 73+64 & & \checkmark& \checkmark \\ 
    & 100.00 & 0+0 && 100.00 & 0+0 && 98.74 & 59+40 & \checkmark & \checkmark  & \checkmark \\ 
    \midrule
    \multirow{6}{*}{\rotatebox{90}{{\textit{\makecell{MS\\($N\approx 25\%$)}}}}}
    & 66.02 & 2497+56   && 99.06 & 2+1 && 94.36 & 136+120 &            &             &            \\
    & 70.46 & 1775+65   && 99.81 & 2+0 && 96.35 & 113+96  & \checkmark &             &            \\
    & 95.37 & 22+56     && 99.10 & 0+2 && 96.88 & 57+105  & \checkmark & \checkmark  &            \\

    & 91.44 & 837+19   && 99.40 & 1+1 && 92.49 & 183+137 & \checkmark &             & \checkmark \\
    & 99.93 & 0+3      && 99.16 & 1+2 && 96.60 & 122+83  &            & \checkmark  & \checkmark \\
    & 100.00 & 0+0     && 99.10 & 0+2 && 97.73 & 67+64   & \checkmark & \checkmark  & \checkmark \\
    \midrule
    \multirow{2}{*}{\rotatebox{90}{{\textit{\makecell{WS\\($0\%$)}}}}} & 74.05 & 438+248 && 93.43 & 4+5 && 91.01 & 201+158 & & \checkmark  & N/A \\
    & 61.49 & 1947+127 && 93.16 & 3+5 && 90.88 & 247+120 & \checkmark & \checkmark  & N/A \\
    
\bottomrule   
\end{tabular}%
}
\end{table*}

\section{Conclusion} \label{sec:conclusion}

In this paper, we presented a deep-learning model for surface-anomaly detection using mixed supervision. We proposed an architecture with two sub-networks: the segmentation sub-network that learns from pixel-level labels and the classification sub-network that learns from weak, image-level labels. Combining both sub-networks allowed for mixing of fully and weakly labeled data to achieve the best result with minimal annotation effort. To accomplish this, we proposed a unified network with end-to-end learning and enabled handling of coarse pixel-level labels. We demonstrated that the proposed model outperforms the existing state-of-the-art anomaly-detection models in both fully supervised mode and in weakly supervised mode. 
The network can be applied in either weakly supervised or fully supervised settings, or in a mixture of both, depending on the availability of annotations. The mixture of both annotation types has resulted in performance competitive to fully supervised mode while requiring significantly fewer and less complex annotations. We have demonstrated this on three existing industrial dataset: DAGM, KolektorSDD and Severstal Steel Defect, and have additionally proposed KolektorSDD2 that represents a real-world industrial problem with several defect types spread over 3000 images. 

Two important conclusions can be drawn from the results. First, using a large number of only weakly labeled data has proven to be more important than using a 10-fold smaller set of fully labelled samples. In cases, such as existing industrial production control where weakly labeled data can be obtained at no cost, it is much less costly to use a large set of weakly labeled data without sacrificing the performance. However, as a second conclusion, we observed that a significant performance boost is often obtained when adding just 5--10\% of fully-labeled data, resulting in almost the same performance as using all fully-labeled data.
This realization can considerably reduce the time and cost of data annotation in many industrial applications.

\section*{Acknowledgments}
This work was in part supported by the ARRS research project J2-9433 (DIVID) and research programme P2-0214. We would also like to thank Kolektor Group d.o.o. for providing images and initial annotations used in KolektorSDD2 dataset.

{\small
\bibliographystyle{ieee}
\bibliography{library}
}

\end{document}